\documentclass[11pt]{article}

\usepackage[final]{acl}

\usepackage{times}
\usepackage{latexsym}

\usepackage[T1]{fontenc}

\usepackage[utf8]{inputenc}

\usepackage{microtype}

\usepackage{inconsolata}

\usepackage{graphicx}
\usepackage{todonotes}
\usepackage{tabularx}

\newcommand{\PlainMedScale}{\textbf{PlainMedScale}}
\newcommand{\nhs}{\textit{NHS}}
\newcommand{\gesundbund}{\textit{Gesund.Bund}}
\newcommand{\apoum}{\textit{ApoUm}}
\newcommand{\apothekenumschau}{\textit{Apotheken Umschau}}
\newcommand{\msd}{\textit{MSD Manuals}}
\newcommand{\msdprof}{\textit{MSD Prof.}}
\newcommand{\msdcons}{\textit{MSD Consumer}}

\usepackage{booktabs}
\usepackage{longtable}

\usepackage{multirow}
\usepackage{makecell}
\usepackage{amsmath}
\usepackage{enumitem}
\usepackage{float}

\usepackage{afterpage}
\usepackage{placeins}

\title{PlainMedScale: A Corpus of Multi-Level Simplified Medical Texts in German and English}

\author{Bruno Brocai \\
  Heidelberg University \\
  German Linguistics \\
  \\\And
  Ilaria Papagno \\
  Heidelberg University  \\
  Translation Studies \\
  \texttt{\{bruno.brocai@gs, ilaria.papagno@iued, ohta@cl\}.uni-heidelberg.de} \\\And
  Mayumi Ohta \\
Heidelberg University  \\
  SFB 1671 \\
}

\begin{document}
\maketitle
\begin{abstract}
We introduce \PlainMedScale, a topic-aligned medical corpus spanning four levels of comprehensibility in German and English, drawn from \msd ~(professional and consumer), \gesundbund, \apothekenumschau ~(\textit{Einfache Sprache}), and the \nhs. The four tiers correspond to distinct communicative functions --- reference, explanation, decision support, and access --- and move beyond the binary expert--lay contrast of prior corpora. In two pilot studies enabled by the alignments, we show that many readability metrics established on two registers fail to generalize across the full gradient, and that a SOTA open-weight LLM prompted for Plain Language still partially preserves the difficulty of its input. Code (\href{https://github.com/GS-Uni-Heidelberg/PlainMedScale}{https://github.com/GS-Uni-Heidelberg/PlainMedScale}) and data (\href{https://doi.org/10.5281/zenodo.21728290}{10.5281/zenodo.21728290}) are made available.
\end{abstract}

\section{Introduction}\label{sec:intro}

Providing high-quality, accessible medical information has been widely acknowledged as a key determinant of enhancing health literacy, defined as individuals' capability to locate, comprehend, and utilize health-related information in their daily lives \citep{sorensen2012, coe2023}. Simplification of medical texts into Plain Language should provide access to relevant medical information for a wider audience, including citizens with low literacy skills, cognitive impairments, or a second language background \citep{inclusionEurope, laban2021}. Recent work on Plain Language (\textit{Einfache Sprache}) guidelines \citep{maass2020, bock2015} supports the translation of specialized content into lay-friendly texts. At the same time, automatic text simplification has advanced considerably \cite{lyu-pergola-2024-society}, offering a potential solution for making specialized knowledge more accessible at scale --- provided suitable corpora.

However, existing German medical text corpora exhibit some limitations. Many present sentence alignment, which enables a detailed analysis of linguistic transformations, but fail to capture content transposition, tending to produce same-length document pairs, without considering the summarization nature of the simplification process, as pointed out by \citet{Gerz:2022}. Most of the existing sources align two registers, a professional version and a single simplified counterpart, thereby reducing medical text simplification to a binary expert-lay contrast. Yet, medical text simplification is not a uniform process but rather a continuum spanning multiple degrees of accessibility. Capturing this gradient requires aligning more than two levels of difficulty. To this end, we extend the English and German texts of the multilingual multiMSDCorpus \cite{horiguchi2025}, with additional lay-oriented resources in both languages. The resulting document-aligned corpus integrates four complementary sources per language, covering different degrees of medical language difficulty. This corpus serves two key purposes: (1) to demonstrate, through a comparison of different steps of linguistic and content transformations across registers, that conventional patient versions no longer meet the required level of acceptability; and (2) to provide four aligned sources at different difficulty levels, offering a more comprehensive picture of the many facets of medical text simplification.
\begin{table*}[t]
\centering
\small
\begin{tabularx}{\textwidth}{lX}
\toprule
\textbf{Source} & \textbf{Text} \\
\midrule
\textbf{MSD Prof.} &
\textit{Etiology of Appendicitis} --- Appendicitis is thought to result from obstruction of the appendiceal lumen, typically by lymphoid hyperplasia but occasionally by a fecalith, foreign body, tumor, or even worms. The obstruction leads to distention, bacterial overgrowth, ischemia, and inflammation. If untreated, necrosis, gangrene, and perforation occur. If the perforation is contained by the omentum, an appendiceal abscess results. \\
\addlinespace
\textbf{MSD Consumer} &
\textit{Causes of Appendicitis} --- The cause of appendicitis is not fully understood. However, in most cases, a blockage inside the appendix probably starts a process. The blockage may be from a small, hard piece of stool (fecalith), a foreign body, tumor, or, rarely, even worms. As a result of the blockage, the appendix becomes inflamed and infected. If inflammation continues without treatment, the appendix can rupture. \\
\addlinespace
\textbf{Gesund.Bund} &
\textit{What causes appendicitis?} --- The appendix is a small protuberance of the cecum that has a narrow lumen (opening). If this lumen is bent or blocked, the appendix swells, interrupting the blood flow. The cells in the wall of the appendix die off and inflammation develops, fostered by bacteria from the intestine. The often purulent inflammation attacks the wall of the appendix, potentially creating a hole. 
\\
\addlinespace
\textbf{NHS} &
\textit{Causes of appendicitis} --- The appendix is a small pouch that's joined to your bowel in the lower right side of your abdomen (tummy). Appendicitis happens when your appendix becomes infected and swollen. This is often caused by something getting stuck in your appendix, such as a small piece of undigested food or hard poo.\\
\bottomrule
\end{tabularx}
\caption{Parallel passages on the etiology of appendicitis across the four sources, illustrating increasing degrees of simplification.}
\label{tab:simplification-example}
\end{table*}

\section{Related Work}
\label{sec:related-work}

For English, Med-EASi \cite{Basu2023} provides short expert–lay text pairs annotated with simplification strategies (elaboration, replacement, insertion, deletion), while JEBS \cite{xia2025} is a larger resource targeting fine-grained lexical simplification of complex biomedical terms in scientific abstracts. At a coarser granularity, the Cochrane corpus \cite{devaraj-etal-2021-paragraph} aligns technical abstracts with plain-language summaries at the paragraph level, and MultiCochrane \cite{joseph-etal-2023-multilingual} extends it to sentence-aligned pairs in English, Spanish, French, and Farsi.

Comparable efforts exist for German, though some remain inaccessible due to data protection restrictions \cite{klaper-etal-2013-building, borchert-etal-2022-ggponc}. General-purpose resources include \citet{battisti-etal-2020-corpus}'s multi-domain corpus for automatic readability assessment and text simplification spanning 92 domains, \citet{jablotschkin-etal-2024-de}'s DE-Lite, which integrates existing corpora with new online data, and \citet{Gerz:2022}'s document-aligned corpus derived from the children's encyclopedia Klexikon. Further corpora include DEplain \cite{stodden-etal-2023-deplain}, which provides document- and sentence-level resources outside the medical domain, and the sentence-aligned dataset of \citet{toborek-etal-2023-new}. Within medicine, the multilingual multiMSDCorpus \cite{horiguchi2025} aligns text across nine languages, including German. However, its patient-oriented version still shows limited readability and a high density of specialized terminology. We therefore propose integrating these text pairs with additional patient-oriented resources offering higher readability and official plain-language resources.

\section{Dataset}\label{sec:dataset}

\subsection{Sources}\label{subsec:sources}
\textit{MSD} \citep{msdmanualsMSDManuals} is a widely used medical resource offering articles in two versions, one for professionals and one for patients, of which we use the German and English editions. \gesundbund~\citep{gesund.bundGesundBund}, provided by the German Federal Ministry of Health, presents conditions and health-related topics in everyday language with little to no technical vocabulary; we again select the German and English editions. The \textit{Einfache Sprache} section of \apothekenumschau~\citep{apothekenumschauApothekenUmschauEinfacher2020} comprises certified plain language texts on diseases and health-related topics in German. At a comparable level in English are the condition descriptions from the official \nhs~website \citep{nhsNHSHealth16Apr20252:51p.m.}, which targets a reading age of 9 to 11. From these sources, the condition sections (but not, e.g., the medication sections) were scraped.

\begin{table}[t]
\centering
\small
\setlength{\tabcolsep}{5pt}
\renewcommand{\arraystretch}{1.15}
\begin{tabular}{cl rrrr}
\toprule
 &  &  & \multicolumn{3}{c}{Aligned w/ $k$ tiers} \\
\cmidrule(lr){4-6}
 & \makecell[l]{Source\\\footnotesize(desc. difficulty)} & $N$ & 1 & 2 & 3 \\
\midrule
\multirow{4}{*}{\rotatebox[origin=c]{90}{German}} & \msdprof & 1594 & 1360 & 157 & 77 \\
 & \msdcons & 1594 & 1360 & 157 & 77 \\
 & \gesundbund & 354 & 42 & 108 & 65 \\
 & \apoum & 180 & 30 & 28 & 81 \\
\midrule
\multirow{4}{*}{\rotatebox[origin=c]{90}{English}} & \msdprof & 1594 & 1286 & 197 & 111 \\
 & \msdcons & 1594 & 1286 & 197 & 111 \\
 & \gesundbund & 354 & 8 & 126 & 47 \\
 & \nhs & 500 & 9 & 166 & 48 \\
\bottomrule
\end{tabular}
\caption{Corpus composition. $N$ = articles per source; columns 1/2/3 = articles aligned with that many of the other three tiers.}
\label{tab:corpus}
\end{table}

\subsection{Instance Alignment}
\label{sec:instance-alignment}
Articles from sources with parallel variants (\gesundbund~across languages, \msd~across difficulty levels) are aligned via site metadata. The remaining articles are routed through shared medical reference vocabularies --- ICD-10 \cite{bfarm-icd10gm-2025}, MeSH \cite{nlm-mesh-2025}, German MeSH \cite{german-mesh-2023}, the Human Disease Ontology (DO)\,\cite{baron2026dokb}, and Pschyrembel \cite{PschyrembelOnline}, and a pair is considered aligned when both articles resolve to the same concept at least once.

\paragraph{Stage~1: Lexical alignment.}
ICD-10 codes embedded in source HTML are extracted at scrape time. To broaden coverage, we index every dictionary entry and its synonyms in SQLite, lemmatized with spaCy\,\citep{honibalSpaCyIndustrialstrengthNatural2020}, and look up each article's canonical title and aliases against the matching-language dictionaries. Matched identifiers propagate across vocabularies via DO cross-references (DOID\,$\leftrightarrow$\,MeSH\,$\leftrightarrow$\,ICD-10CM).

\paragraph{Stage~2: Embedding-based alignment.}
Lexical matching is high-recall but noisy. We embed each article's first paragraph and every dictionary entry with Qwen3-Embedding-8B\,\cite{qwen3embedding} and (i)~re-rank Stage~1 candidates by L2 distance, retaining the closest; (ii)~for articles without any Stage~1 hit, run top-$k$ retrieval against each dictionary, accepting a match only if a reverse full-text lookup of its title resolves back to the same article. Two articles are linked whenever a chain of shared concept IDs connects them, and we rank such links by the embedding distance between the articles.

\begin{table}
\centering
\small
\setlength{\tabcolsep}{4pt}
\renewcommand{\arraystretch}{1.1}
\begin{tabular}{l c cccc}
\toprule
 & & \multicolumn{4}{c}{Per-class accuracy (\%)} \\
\cmidrule(lr){3-6}
Model (reasoning) & Overall & corr. & spec./gen. & inc. \\
\midrule
\textbf{GPT-5.4-mini (none)}        & 80.0 & 77.1 & 41.4 / 66.7 & 90.2 \\
GPT-5.4-mini (low)         & 80.1 & 71.8 & 60.9 / 75.0 & 88.1 \\
GPT-5.4 (none)             & 80.2 & 69.4 & 64.8 / 75.0 & 88.4 \\
Claude Haiku 4.5 (none)    & 77.3 & 73.9 & 48.4 / 66.7 & 85.5 \\
Qwen3-30B-Instruct (-) & 75.3 & 88.2 & 21.9 /  8.3 & 83.2 \\
\midrule
$n$                                 & 956  & 245  & 128 / 12   & 571 \\
\bottomrule
\end{tabular}
\caption{Four-class LLM-judge accuracy on the 956-pair manual gold
set.}
\label{tab:llm-judge-models}
\end{table}

\paragraph{Stage 3: LLM-based filtering.}
Two annotators with medical backgrounds labeled Apotheken-Umschau-anchored pairs as \texttt{correct} (the same topic), \texttt{specialization} (source is narrower than \apothekenumschau), \texttt{generalization} (broader), and \texttt{incorrect} (unrelated), resulting in $956$ pairs. The annotation showed that many pairs are partial topic matches (e.g., \textit{Paralysis–Vocal Fold Paralysis}), and that Stage 1+2 precision is low overall.

We evaluate different candidate models (Table~\ref{tab:llm-judge-models}) to be used in the follow-up filtering step. We observe a tradeoff: larger or reasoning-enabled models do better on the harder \texttt{specialization/generalization} distinction, but this comes at the cost of accuracy on the \texttt{correct/incorrect} classes --- the two that matter most for corpus quality.
GPT-5.4-mini flags the most genuine \texttt{incorrect} alignments while trailing only Qwen3 on \texttt{correct}; Qwen3's advantage, however, comes with weak \texttt{incorrect} detection. We therefore pick GPT-5.4-mini as our judge for the full data.


We noticed the original labels sometimes collapsed a cluster of related titles into one judgment, so a pair that was actually a distinct match could inherit a label that did not fit. To quantify this, a second reviewer re-annotated 100 pairs, yielding 92\,\% precision on \texttt{correct} (95\,\% Wilson CI 67--99) and $\geq$\,88\,\% on the other classes. Both numbers are biased: the 92\,\% likely overestimates true agreement, since the reviewer saw each LLM label and was asked to override only disagreements (anchoring bias toward the LLM's judgment); the original reviewer accuracy likely underestimates it, due to the issue described above. True precision most likely falls between the two, a level we consider acceptable.
\section{Text Assessment}

\subsection{Readability Indices}

\begin{table*}[!t]
\centering
\small
\begin{tabular}{lccccccc}
\toprule
Metric & Apotheken Umschau &  & Gesund.Bund &  & MSD Cons. &  & MSD Prof. \\
\midrule
\textbf{Wiener Sachtextformel} & 6.80 $\pm$ 1.05 & * & 10.29 $\pm$ 1.03 & * & 10.76 $\pm$ 1.24 & * & 12.23 $\pm$ 1.52 \\
Noun ratio & 0.30 $\pm$ 0.02 & \textdagger & 0.27 $\pm$ 0.02 & * & 0.33 $\pm$ 0.03 & * & 0.36 $\pm$ 0.05 \\
\underline{Adjective ratio} & 0.04 $\pm$ 0.02 & * & 0.06 $\pm$ 0.02 &  & 0.07 $\pm$ 0.02 & * & 0.09 $\pm$ 0.02 \\
Function word ratio & 0.68 $\pm$ 0.02 & \textdagger & 0.69 $\pm$ 0.02 & * & 0.68 $\pm$ 0.03 & \textdagger & 0.71 $\pm$ 0.04 \\
\underline{Numbers ratio} & 0.00 $\pm$ 0.01 & * & 0.01 $\pm$ 0.00 &  & 0.01 $\pm$ 0.01 & * & 0.02 $\pm$ 0.02 \\
\underline{Negations ratio} & 0.01 $\pm$ 0.00 &  & 0.01 $\pm$ 0.00 & * & 0.01 $\pm$ 0.00 & * & 0.01 $\pm$ 0.00 \\
Grammar frequency & 0.85 $\pm$ 0.05 & \textdagger & 0.95 $\pm$ 0.04 & * & 0.89 $\pm$ 0.08 & * & 0.87 $\pm$ 0.12 \\
\textbf{Term frequency (PMW)} & 70.3 $\pm$ 15.3 & * & 54.1 $\pm$ 10.4 & * & 44.3 $\pm$ 12.0 & * & 34.4 $\pm$ 11.5 \\
Avg. lexical chain length & 0.01 $\pm$ 0.00 & * & 0.00 $\pm$ 0.00 & * & 0.01 $\pm$ 0.00 & \textdagger & 0.01 $\pm$ 0.01 \\
Avg. lexical cross chains & 0.49 $\pm$ 0.18 & * & 0.82 $\pm$ 0.32 & * & 1.00 $\pm$ 0.81 & \textdagger & 0.89 $\pm$ 0.83 \\
\textbf{Jargon density} & 0.043 $\pm$ 0.022 & * & 0.074 $\pm$ 0.025 & * & 0.099 $\pm$ 0.031 & * & 0.148 $\pm$ 0.041 \\
\bottomrule
\end{tabular}
\caption{Readability metrics (DE) positively evaluated by \citet{scholzEvaluatingReadabilityMetrics2025}, mean $\pm$ std across articles per source. Markers between adjacent low$\rightarrow$high-readability data sources indicate the significant shift in the expected direction (*), or in the opposite direction (\textdagger) (one-sided Wilcoxon signed-rank on aligned article pairs, $\alpha{=}0.05$, Bonferroni-corrected). Because the test was only done on aligned pairs, full-corpus mean and significance might not agree. Bold/underlined: metrics that are fully/partially significant only in the expected direction.}
\label{tab:readability-de}
\end{table*}

We apply a subset of the readability metrics from \citet{scholzEvaluatingReadabilityMetrics2025}, evaluated there on German and English Plain Language summaries; we refer to that paper for the exact equations and NLP pipeline. We report two additional metrics: since Flesch-Kincaid Grade Level is not suited for German, we report the Wiener Sachtextformel \cite{bamberger1984lesen} for that language, and we estimate the frequency of medical jargon via a corpus-based keyness approach (full procedure in appendix \ref{app:jargon}). Table \ref{tab:readability-de} reports the German results (English in appendix \ref{app:eng-readability}).

The aligned nature of our corpus lets us test the statistical significance of readability metrics across difficulty levels while minimizing the topic confounder, using a Wilcoxon signed-rank test on each pairwise difficulty step. Surprisingly, most readability metrics show no clear slope across the four sources: the noun ratio, for instance, is lowest for \gesundbund~and significantly higher for \apothekenumschau. The only metrics that hold up fully for German are the Wiener Sachtextformel, the adjective and negation ratios, and the related term Frequency and jargon density measures.

Overall, these findings point to a problem in evaluating readability metrics on only two corpora: any score difference observed there may hold at that specific difficulty step without generalizing across the full readability range.

\begin{figure}
    \centering
    \includegraphics[width=1\linewidth]{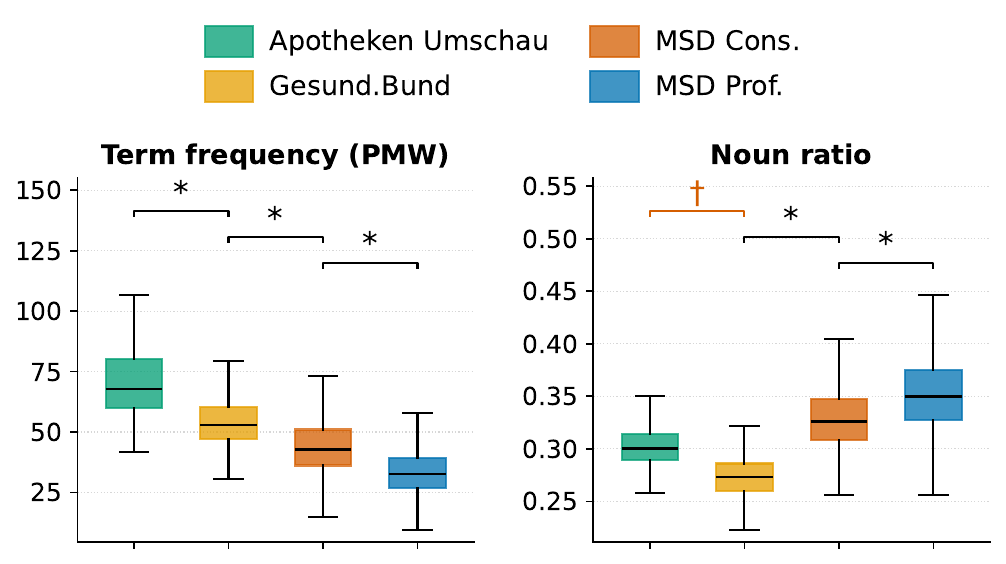}
    \caption{Two metrics from Table \ref{tab:readability-de}: term frequency shifts as expected at every step, noun ratio does not (†).}
\end{figure}

\begin{figure*}[h]
    \centering
    \includegraphics[width=1\linewidth]{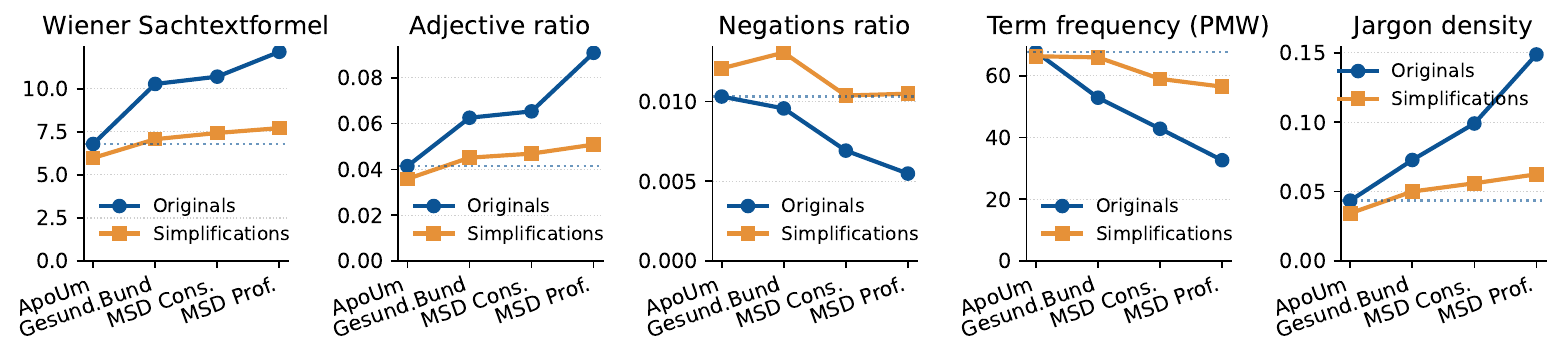}
    \caption{Input complexity bleed-through in simplification outputs for German. 
    On most monotonic metrics, shown here, the simplification reaches roughly Plain Language (\apothekenumschau) level. Note that \apothekenumschau~was simplified for completeness, but given it already has a plain language level, this task is redundant.}
    \label{fig:simp-bleedthrough}
\end{figure*}

\subsection{Content Analysis}

The sources in our dataset differ not only in linguistic complexity, but also in their communicative function (cf. \citealp{jakobson1960}).
Through a qualitative reading of a sample of aligned articles, we identified four shared functions, each of which was more prominent in a single tier: a reference function (\msdprof), presenting information flatly and systematically; an explanatory function (\msdcons), conveying pathophysiological accounts to lay readers through narrative framing; a decision-support function (\gesundbund), providing evidence-based statements and quality-of-life considerations to inform autonomous patient decisions; and an access function (\apothekenumschau~Plain Language, \nhs), offering action-oriented, low-threshold information organized around readers' likely concerns. This functional differentiation explains why certain information appears in some sources but not in others, and provides a normative criterion for what should be preserved during simplification.

Plain language must not only remove linguistic obstacles --- such as complex syntax and jargon --- but also reshape content. Defining a threshold of content completeness is challenging, since omission is intrinsic to simplification and its adequacy is determined by its function. Comparing different levels of human simplification helps identify what should be retained. Patient-oriented versions of MSD texts still contain substantial medical jargon --- uncommon anatomical terms, adjectives, and verbs --- often left undefined or explained only later in the text, while illustrations rely on classical terminology. Their communicative function remains referential: information is presented without a clear hierarchy of relevance and offers little practical guidance. \gesundbund~and \apothekenumschau, by contrast, indicate when a symptom requires medical intervention or when a given procedure is routine, guiding readers' judgment without demanding specialized knowledge. This is particularly relevant for German, whose medical lexicon distinguishes sharply between German-rooted terms and Greek/Latin expressions typical of medical reports (e.g. ``Blinddarmentz\"{u}ndung'' $\leftrightarrow$ ``Appendizitis''); patients cannot be assumed to recognize these as referring to the same condition. \apothekenumschau~bridges this gap with clarifications such as ``Der Arzt sagt dazu: Appendizitis.'' [The doctor calls it: Appendicitis]. \nhs~and \apothekenumschau~further reduce terminological precision, yielding shorter, more linear descriptions that preserve the core information while minimizing the need for explanation.

\subsection{Simplification Experiment}

We probe whether source difficulty leaves a residue in LLM simplifications. Using the cross-source aligned passages, we prompt
Qwen3-30B-A3B-Instruct-2507 \cite{qwen3technicalreport} to rewrite each text at Plain Language / Einfache Sprache level, then check whether the outputs still track source difficulty on the strictly monotonic readability metrics. A perfect simplifier would erase the ordering; instead, every testable metric except the negation ratio retains it (Figure \ref{fig:simp-bleedthrough}). 

However, this bleed-through is weak on average. In addition, the model mostly achieves exactly the metric of the human original Plain Language text. In terms of these metrics, then, simplification works relatively well.


\section{Conclusion}

We introduce \PlainMedScale, a topic-aligned medical corpus spanning
four comprehensibility levels in German and English. Moving beyond the
binary expert--lay contrast, the corpus exposes simplification as a
continuum of communicative functions: reference (\msdprof),
explanation (\msdcons), decision support (\gesundbund), and access
(\apothekenumschau~/ \nhs).

Two findings stand out. Most readability metrics established on two
registers do not generalize across the full gradient: in German, only
the Wiener Sachtextformel, the adjective and negation ratios, and the
corpus-based term-frequency and jargon-density measures shift
monotonically across all steps. And when asked for Plain
Language, a competitive open-weight LLM partially preserves input
difficulty, though it roughly reaches the target level. The corpus thus
supports stress-testing readability metrics, training, and evaluating systems against multiple targets, and --- via the
bilingual alignments of \msd{} and \gesundbund{} --- studying simplification
alongside translation.

\subsection*{Limitations} 
Authorship and translation direction vary across sources (\gesundbund~authored in German, \msd~in English), so cross-lingual comparisons conflate the two. The lowest-difficulty tier is functionally but not editorially parallel: \apothekenumschau~follows German Plain Language conventions, while \nhs~uses its own house style. Finally, our alignment pipeline yields no measurable recall figure, since pairs sharing no vocabulary anchor are silently missed. And validation is anchored on Apotheken Umschau pairs only, and we assume the resulting precision transfers to the rest of the corpus.

\section*{Acknowledgments}
The authors acknowledge support by the state of Baden-W\"{u}rttemberg through bwHPC and the German Research Foundation (DFG) through grant INST 35/1597-1 FUGG.

\bibliography{bib}

\appendix

\section{Jargon density}
\label{app:jargon}
We tokenize each article with SoMaJo \cite{Proisl_Uhrig_EmpiriST:2016} and classify a token as \emph{jargon} when its keyness in medical research texts against a general-language reference corpus exceeds an odds ratio of $20$ \emph{and} its relative frequency in that reference corpus is at most $0.001$ (the latter filter removes common-but-keyed terms such as \emph{Patient}). Out-of-vocabulary tokens --- i.e.,\ tokens absent from the keyness table --- are also counted as jargon. The reported metric is \textit{jargon tokens per all tokens}, $N_{\text{jargon}} / N_{\text{tokens}}$. Keyness is precomputed per language: we use DRKS-abstracts \cite{DRKSDeutschesRegister} for German and a sample of PubMed \cite{PubMed} articles for English medical language; and we use Leipzig Newspaper Corpora \cite{goldhahnBuildingLargeMonolingual} as general-language reference. Because the two corpora differ in size and register, and because SoMaJo keeps German nominal compounds as single tokens, within-language source comparisons are meaningful but cross-language differences are not.

\section{Readability scores}

Refer to \citet{scholzEvaluatingReadabilityMetrics2025} for the exact definition of these metrics.

\subsection{Abbreviations}

\begin{description}[style=unboxed, leftmargin=0pt, font=\normalfont\bfseries]
    \item[FKGL] Flesch-Kincaid grade level
    \item[CLS score] Next-sentence prediction score: how confident BERT is that one sentence naturally follows the previous one (based on BERT's \texttt{[CLS]} token)
    \item[Term Frequency] Mean frequency of content words in general-language corpora. Displayed as \textbf{PMW} (per million words) for better readability
    \item[Grammar Frequency] How many different sentence structures occur in a text, relative to the number of sentences — a text with more repeated structures gets a lower score
    \item[Lexical chain] A noun that is repeated across a text
    \item[NP] Noun phrase
    \item[Edit distance] How much a sentence's structure changes compared to the sentence before it
    \item[NP complexity] Noun phrase complexity
    \item[Fill Mask probability]  How easily BERT can restore randomly masked words, measured as the true word's normalized rank among
    all vocabulary candidates. Lower values mean better prediction
\end{description}

Fill-mask probability and average CLS score are computed using BERT\footnote{\href{https://huggingface.co/google-bert/bert-base-uncased}{https://huggingface.co/google-bert/bert-base-uncased}} for English texts and German BERT\footnote{\href{https://huggingface.co/google-bert/bert-base-german-cased}{https://huggingface.co/google-bert/bert-base-german-cased}} for German texts. Average sentence perplexity is computed using GPT-2\footnote{\href{https://huggingface.co/openai-community/gpt2}{https://huggingface.co/openai-community/gpt2}} for English texts and German GPT-2\footnote{\href{https://huggingface.co/dbmdz/german-gpt2}{https://huggingface.co/dbmdz/german-gpt2}} for German texts.

\subsection{Full German readability scores}
\label{app:german-readability}
\begin{figure*}[t]
\noindent
\centering
\includegraphics[width=\linewidth]{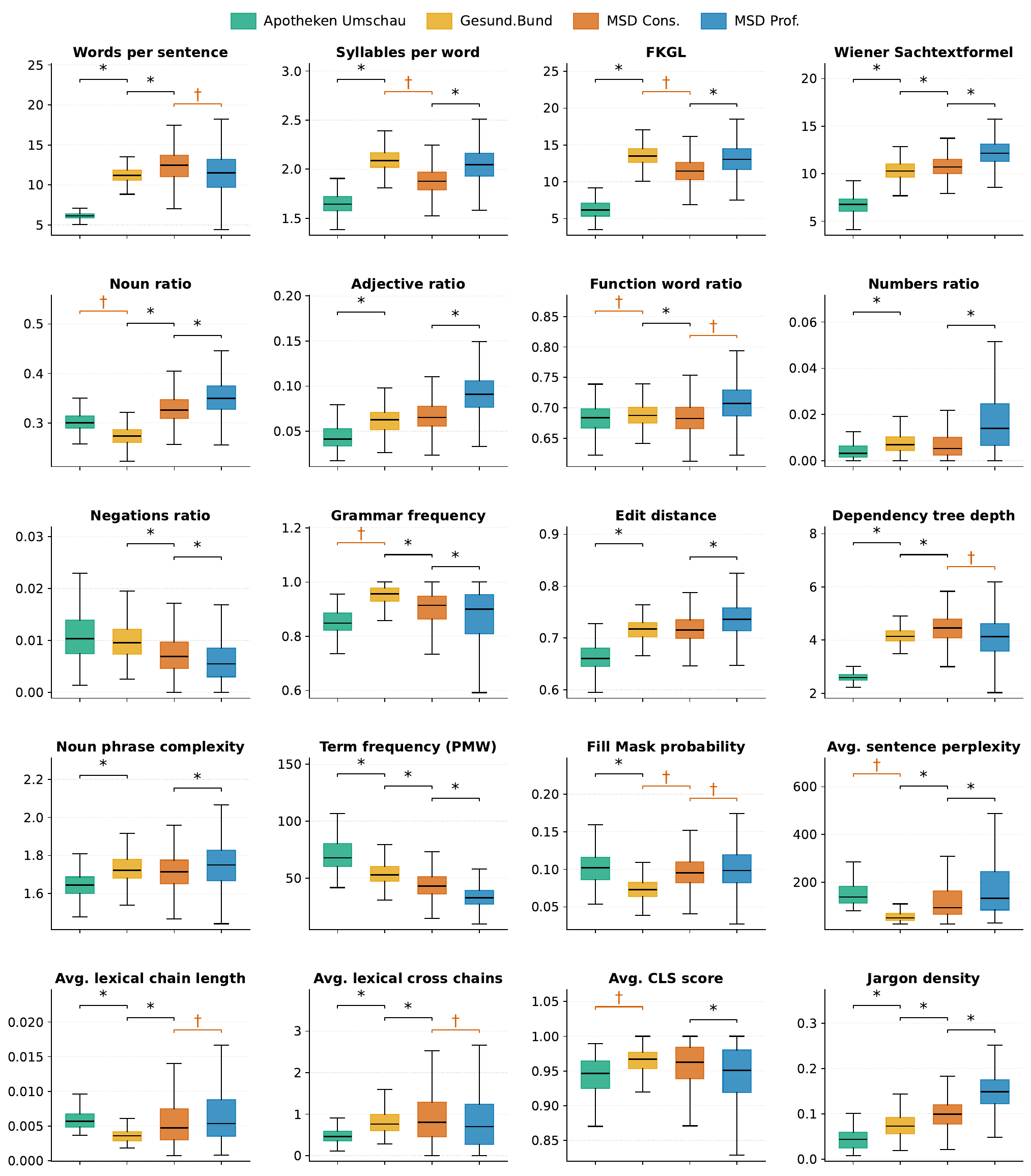}
\caption{German readability metrics}
\label{fig:readability-de}
\end{figure*}

See Table \ref{tab:readability-de-leftovers} and Figure \ref{fig:readability-de}.

\begin{table*}[!b]
\centering
\small
\begin{tabular}{lccccccc}
\toprule
Metric & Apotheken Umschau &  & Gesund.Bund &  & MSD Cons. &  & MSD Prof. \\
\midrule
Words per sentence & 6.11 $\pm$ 0.45 & * & 11.23 $\pm$ 0.95 & * & 12.44 $\pm$ 2.23 & \textdagger & 11.46 $\pm$ 2.90 \\
Syllables per word & 1.65 $\pm$ 0.11 & * & 2.10 $\pm$ 0.11 & \textdagger & 1.88 $\pm$ 0.15 & * & 2.05 $\pm$ 0.20 \\
FKGL & 6.27 $\pm$ 1.35 & * & 13.53 $\pm$ 1.28 & \textdagger & 11.49 $\pm$ 1.79 & * & 13.11 $\pm$ 2.29 \\
\underline{Edit distance} & 0.66 $\pm$ 0.03 & * & 0.72 $\pm$ 0.02 &  & 0.72 $\pm$ 0.03 & * & 0.73 $\pm$ 0.04 \\
Dependency tree depth & 2.59 $\pm$ 0.17 & * & 4.15 $\pm$ 0.27 & * & 4.42 $\pm$ 0.59 & \textdagger & 4.09 $\pm$ 0.82 \\
\underline{Noun phrase complexity} & 1.65 $\pm$ 0.08 & * & 1.73 $\pm$ 0.08 &  & 1.72 $\pm$ 0.10 & * & 1.75 $\pm$ 0.13 \\
Fill Mask probability & 0.102 $\pm$ 0.022 & * & 0.074 $\pm$ 0.014 & \textdagger & 0.097 $\pm$ 0.026 & \textdagger & 0.102 $\pm$ 0.033 \\
Avg. sentence perplexity & 301 $\pm$ 1172 & \textdagger & 68 $\pm$ 61 & * & 288 $\pm$ 1234 & * & 413 $\pm$ 2820 \\
Avg. CLS score & 0.943 $\pm$ 0.027 & \textdagger & 0.963 $\pm$ 0.021 &  & 0.957 $\pm$ 0.039 & * & 0.942 $\pm$ 0.056 \\
\bottomrule
\end{tabular}
\caption{German readability metrics not in the main article table (i.e., all metrics deemed problematic for German by \citet{scholzEvaluatingReadabilityMetrics2025}), mean $\pm$ std across articles per source. Markers between adjacent low$\rightarrow$high-readability data sources indicate the significant shift in the expected direction (*), or in the opposite direction (\textdagger) (one-sided Wilcoxon signed-rank on aligned article pairs, $\alpha{=}0.05$, Bonferroni-corrected). Because the test was only done on aligned pairs, full-corpus mean and significance might not agree. Bold/underlined: metrics that are fully/partially significant only in the expected direction.}
\label{tab:readability-de-leftovers}
\end{table*}

\subsection{Full English readability scores}
\label{app:eng-readability}
\begin{figure*}[t]
\centering
\includegraphics[width=\linewidth]{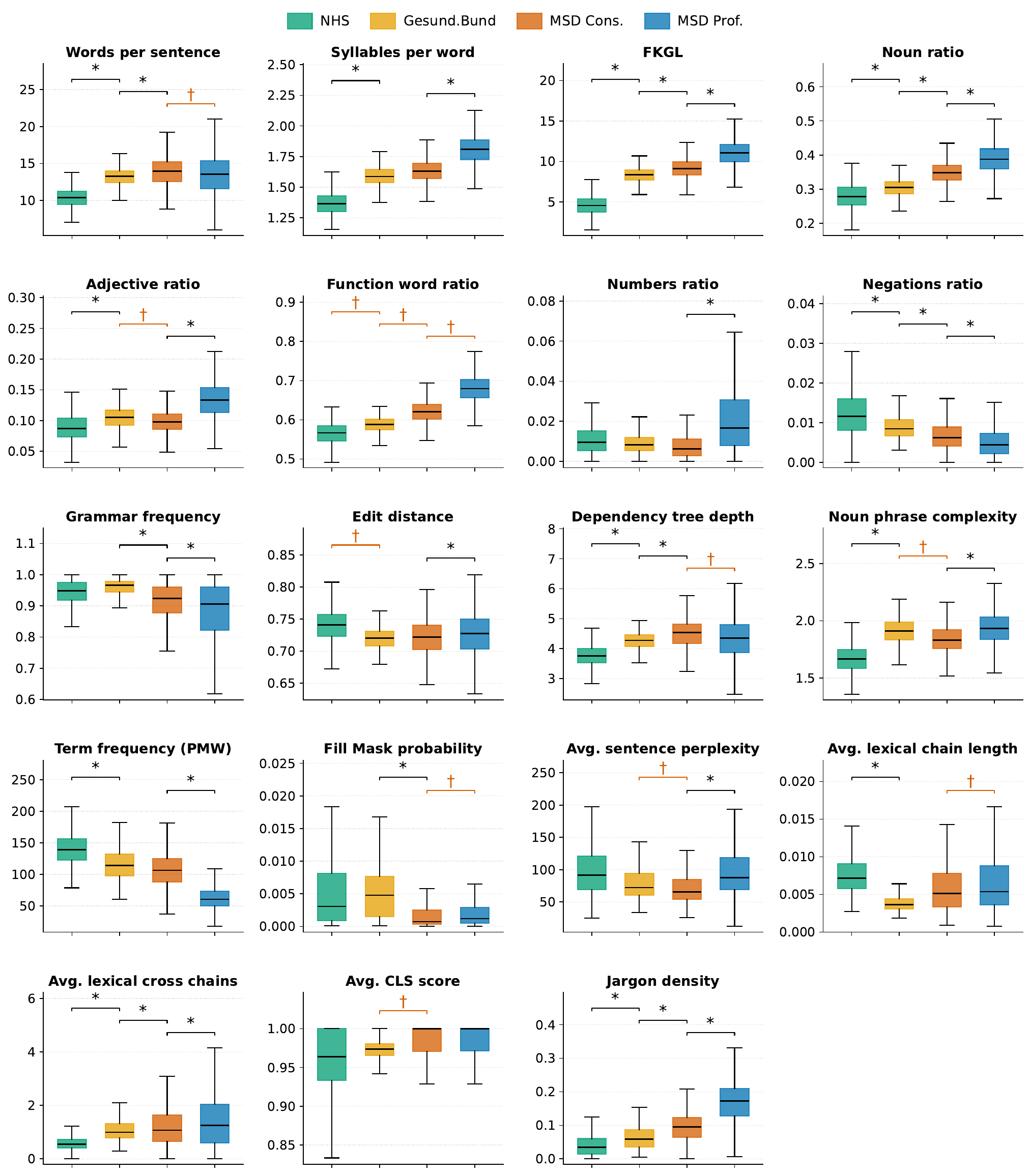}
\caption{English readability metrics}
\label{fig:readability-en}
\end{figure*}

See Table \ref{tab:readability-en} and Figure \ref{fig:readability-en}.

\begin{table*}[b]
\centering
\small
\begin{tabular}{lccccccc}
\toprule
Metric & NHS &  & Gesund.Bund &  & MSD Cons. &  & MSD Prof. \\
\midrule
Words per sentence & 10.43 $\pm$ 1.41 & * & 13.22 $\pm$ 1.22 & * & 13.98 $\pm$ 2.31 & \textdagger & 13.51 $\pm$ 3.13 \\
\underline{Syllables per word} & 1.37 $\pm$ 0.09 & * & 1.59 $\pm$ 0.08 &  & 1.64 $\pm$ 0.10 & * & 1.81 $\pm$ 0.13 \\
\textbf{FKGL} & 4.62 $\pm$ 1.23 & * & 8.30 $\pm$ 0.93 & * & 9.16 $\pm$ 1.38 & * & 11.04 $\pm$ 1.76 \\
\textbf{Noun ratio} & 0.28 $\pm$ 0.04 & * & 0.30 $\pm$ 0.03 & * & 0.35 $\pm$ 0.04 & * & 0.39 $\pm$ 0.05 \\
Adjective ratio & 0.09 $\pm$ 0.02 & * & 0.11 $\pm$ 0.02 & \textdagger & 0.10 $\pm$ 0.02 & * & 0.13 $\pm$ 0.03 \\
Function word ratio & 0.57 $\pm$ 0.03 & \textdagger & 0.59 $\pm$ 0.02 & \textdagger & 0.62 $\pm$ 0.03 & \textdagger & 0.68 $\pm$ 0.04 \\
\underline{Numbers ratio} & 0.01 $\pm$ 0.01 &  & 0.01 $\pm$ 0.01 &  & 0.01 $\pm$ 0.01 & * & 0.02 $\pm$ 0.02 \\
\textbf{Negations ratio} & 0.01 $\pm$ 0.01 & * & 0.01 $\pm$ 0.00 & * & 0.01 $\pm$ 0.00 & * & 0.01 $\pm$ 0.00 \\
\underline{Grammar frequency} & 0.94 $\pm$ 0.04 &  & 0.96 $\pm$ 0.03 & * & 0.91 $\pm$ 0.07 & * & 0.88 $\pm$ 0.10 \\
Edit distance & 0.74 $\pm$ 0.03 & \textdagger & 0.72 $\pm$ 0.02 &  & 0.72 $\pm$ 0.03 & * & 0.73 $\pm$ 0.04 \\
Dependency tree depth & 3.78 $\pm$ 0.35 & * & 4.27 $\pm$ 0.32 & * & 4.51 $\pm$ 0.55 & \textdagger & 4.34 $\pm$ 0.74 \\
NP complexity & 1.67 $\pm$ 0.13 & * & 1.91 $\pm$ 0.11 & \textdagger & 1.84 $\pm$ 0.13 & * & 1.94 $\pm$ 0.16 \\
\underline{Term frequency (PMW)} & 141.2 $\pm$ 28.9 & * & 116.7 $\pm$ 27.0 &  & 108.2 $\pm$ 30.2 & * & 64.0 $\pm$ 21.5 \\
Fill Mask probability & 0.005 $\pm$ 0.006 &  & 0.006 $\pm$ 0.005 & * & 0.002 $\pm$ 0.004 & \textdagger & 0.003 $\pm$ 0.004 \\
Avg. sentence perplexity & 123 $\pm$ 151 &  & 89 $\pm$ 64 & \textdagger & 93 $\pm$ 226 & * & 141 $\pm$ 501 \\
Avg. lexical chain length & 0.01 $\pm$ 0.00 & * & 0.00 $\pm$ 0.00 &  & 0.01 $\pm$ 0.00 & \textdagger & 0.01 $\pm$ 0.01 \\
\textbf{Avg. lexical cross chains} & 0.58 $\pm$ 0.27 & * & 1.09 $\pm$ 0.41 & * & 1.24 $\pm$ 0.81 & * & 1.46 $\pm$ 1.12 \\
Avg. CLS score & 0.958 $\pm$ 0.041 &  & 0.971 $\pm$ 0.015 & \textdagger & 0.980 $\pm$ 0.032 &  & 0.980 $\pm$ 0.038 \\
\textbf{Jargon density} & 0.040 $\pm$ 0.029 & * & 0.063 $\pm$ 0.032 & * & 0.094 $\pm$ 0.042 & * & 0.167 $\pm$ 0.063 \\
\bottomrule
\end{tabular}
\caption{English readability metrics, mean $\pm$ std across articles per source. Markers between adjacent low$\rightarrow$high-readability data sources indicate the significant shift in the expected direction (*), or in the opposite direction (\textdagger) (one-sided Wilcoxon signed-rank on aligned article pairs, $\alpha{=}0.05$, Bonferroni-corrected). Because the test was only done on aligned pairs, full-corpus mean and significance might not agree. Bold/underlined: metrics that are fully/partially significant only in the expected direction.}
\label{tab:readability-en}
\end{table*}

\end{document}